\documentclass[conference]{IEEEtran}

\usepackage{fancyhdr}
\usepackage[normalem]{ulem}
\usepackage[hyphens]{url}
\usepackage{hyperref}
\usepackage{microtype}
\usepackage{amsmath}
\usepackage{graphicx}
\usepackage{epsfig}
\usepackage{subfigure}
\usepackage{multirow}
\usepackage{array}
\usepackage{enumitem}
\usepackage{cleveref}
\usepackage[numbers,sort&compress]{natbib}
\usepackage{tablefootnote}
\usepackage{url}
\usepackage{cleveref}

\begin{document}
%
\title{Hardware for Machine Learning: \\Challenges and Opportunities}

\author{\IEEEauthorblockN{Vivienne Sze, Yu-Hsin Chen, Joel Emer, Amr Suleiman, Zhengdong Zhang}
\IEEEauthorblockA{
Massachusetts Institute of Technology\\
Cambridge, MA 02139}

}
 

%

\IEEEspecialpapernotice{(Invited Paper)}

\maketitle

\begin{abstract}
Machine learning plays a critical role in extracting meaningful information out of the zetabytes of sensor data collected every day. For some applications, the goal is to analyze and understand the data to identify trends (e.g., surveillance, portable/wearable electronics); in other applications, the goal is to take immediate action based the data (e.g., robotics/drones, self-driving cars, smart Internet of Things). For many of these applications, local embedded processing near the sensor is preferred over the cloud due to privacy or latency concerns, or limitations in the communication bandwidth. However, at the sensor there are often stringent constraints on energy consumption and cost in addition to throughput and accuracy requirements. Furthermore, flexibility is often required such that the processing can be adapted for different applications or environments (e.g., update the weights and model in the classifier). In many applications, machine learning often involves transforming the input data into a higher dimensional space, which, along with programmable weights, increases data movement and consequently energy consumption. In this paper, we will discuss how these challenges can be addressed at various levels of hardware design ranging from architecture, hardware-friendly algorithms, mixed-signal circuits, and advanced technologies (including memories and sensors).  
\end{abstract}


%
\IEEEpeerreviewmaketitle

\section{Introduction}
This is the era of \emph{big data}.  More data has been created in the past two years than the entire history of the human race~\cite{forbes}.  This is primarily driven by the exponential increase in the use of sensors (10 billion per year in 2013, expected to reach 1 trillion by 2020~\cite{sensors}) and connected devices (6.4 billion in 2016, expected to reach 20.8 billion by 2020~\cite{connected}). These sensors and devices generate hundreds of zetabytes ($10^{21}$ bytes) of data per year ---  petabytes ($10^{15}$ bytes) per second~\cite{ciscoGCI}.

\emph{Machine learning} is needed to extract meaningful, and ideally actionable, information from this data.  A significant amount of computation is required to analyze this data, which often happens in the cloud. However, given the sheer volume and rate at which data is being generated, and the high energy cost of communication and often limited bandwidth, there is an increasing need to perform the analysis locally near the sensor rather than sending the raw data to the cloud. Embedding machine learning at the edge also addresses important concerns related to privacy, latency and security.

\section{Applications}
Many applications can benefit from embedded machine learning ranging from multimedia to medical space. We will provide a few examples of areas that researchers have investigated; however, this paper will primarily focus on computer vision, specifically image classification, as a driving example.

\subsection{Computer Vision}
Video is arguably the biggest of the big data.  It accounts for over 70\% of today's Internet traffic~\cite{ciscoVNI}.  For instance, over 800 million hours of video is collected daily worldwide for video surveillance~\cite{video_surveillance}.  In many applications (e.g., measuring wait times in stores, traffic patterns), it would be desirable to use computer vision to extract the meaningful information from the video right at the image sensor rather than in the cloud to reduce the communication cost. For other applications such as autonomous vehicles, drone navigation and robotics, local processing is desired since the latency and security risk of relying on the cloud are too high. However, video involves a large amount of data, which is computationally complex to process; thus, low cost hardware to analyze video is challenging yet critical to enabling these applications.  

\begin{figure}
    \begin{center}
        \includegraphics[width=0.9\linewidth]{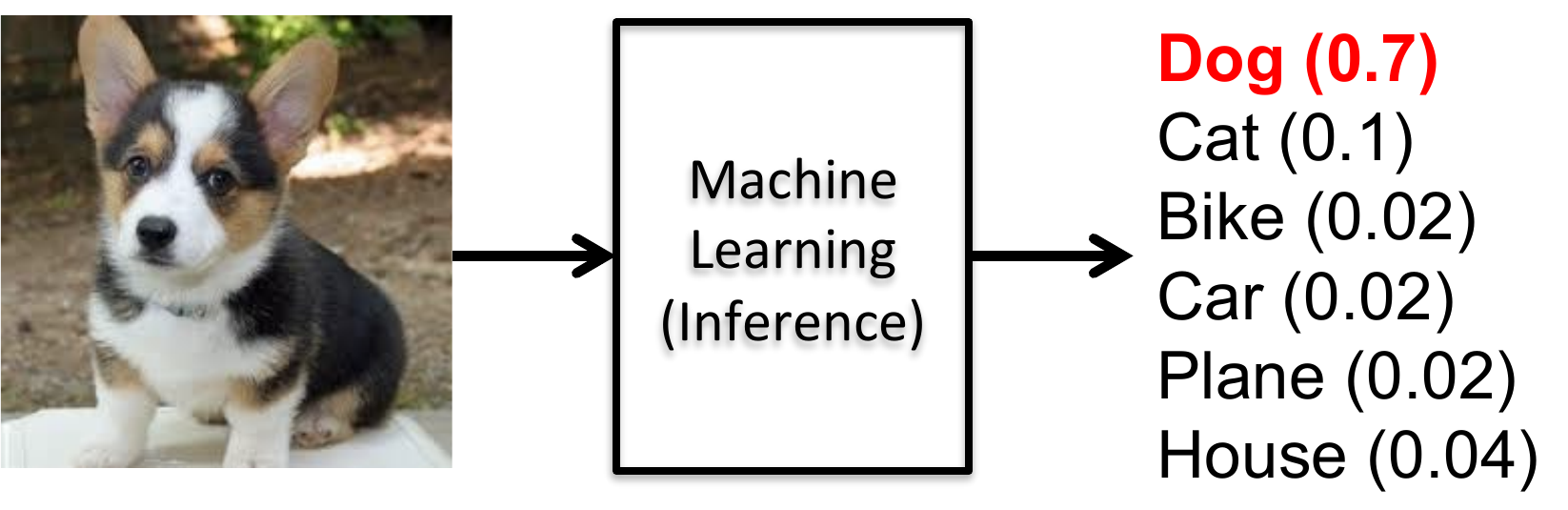}
                \vspace{-5pt}
        \caption{    Image classification.
                }
        \vspace{-15pt}                
        \label{fig:image_classification}
    \end{center}
\end{figure}

In computer vision, there are many different artificial intelligence (AI) tasks~\cite{szeliski2010computer}. In this paper, we focus on image classification (Fig.~\ref{fig:image_classification}), where the entire image is provided and the task is to determine which class of objects is in the image. 

\subsection{Speech Recognition} 
Speech recognition has significantly improved our ability to interact with  electronic devices, such as smartphones.  While currently most of the processing for applications such as Apple Siri and Amazon Alexa voice services is in the cloud, it is desirable to perform the recognition on the device itself to reduce latency and dependence on connectivity, and to increase privacy.  Speech recognition is the first step before many other AI tasks such as machine translation, natural language processing, etc. Low power hardware for speech recognition is explored in~\cite{price20156,yazdaniultra}.

\subsection{Medical} 
There is a strong clinical need to be able to monitor patients and to collect long-term data to help either detect/diagnose various diseases or monitor treatment.  For instance, constant monitoring of ECG or EEG signals would be helpful in identifying cardiovascular diseases or detecting the onset of a seizure for epilepsy patients, respectively. In many cases, the devices are either wearable or implantable, and thus the energy consumption must be kept to a minimum.  Using embedded machine learning to extract meaningful physiological signal and process it locally is explored in~\cite{verma2009micro,chen20101,lee2013low}.

\section{Machine Learning Basics}

Machine learning is a form of artificial intelligence (AI) that can perform a task without being specifically programmed. Instead, it \emph{learns} from previous examples of the given task during a process called \emph{training}.  After learning, the task is performed on new data through a process called \emph{inference}. Machine learning is particularly useful for applications where the data is difficult to model analytically.

Training involves learning a set of weights from a dataset. When the data is labelled, it is referred to as supervised learning, which is currently the most widely-used approach. Inference involves performing a given task using the learned weights (e.g., classify an object in an image)\footnote{Machine learning can be used in a discriminative or generative manner. This paper focuses on the discriminative use.}. In many cases, training is done in the cloud.  Inference can also happen in the cloud; however, as previously discussed, for certain applications this is not desirable from the standpoint of communication, latency and privacy. Instead it is preferred that the inference occur locally on a device near the sensor.  In these cases, the trained weights are downloaded from the cloud and stored in the device. Thus, the device needs be programmable in order to support a reasonable range of tasks.

\begin{figure}
    \begin{center}
        \includegraphics[width=0.9\linewidth]{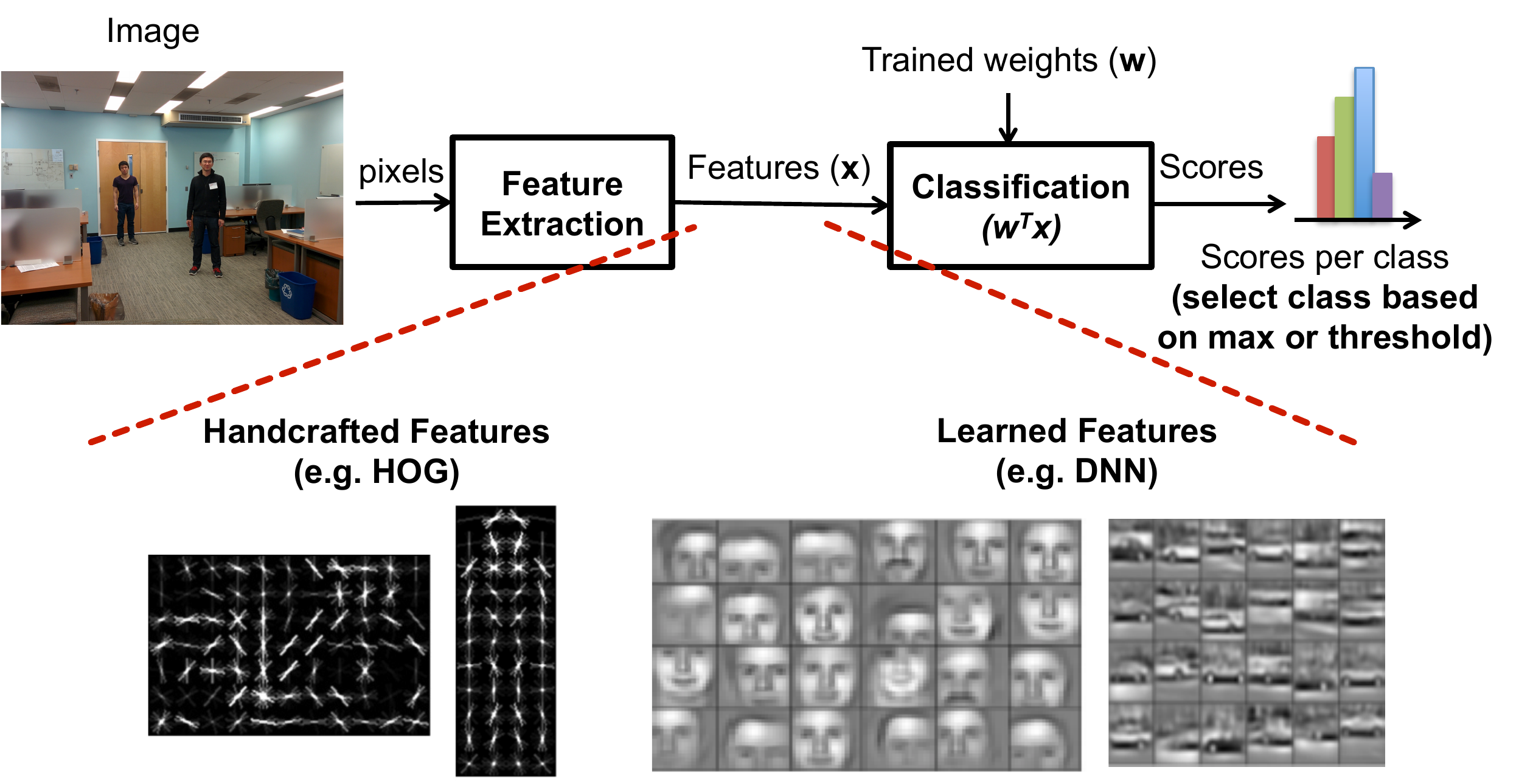}
        \caption{    Inference pipeline.
                }
        \vspace{-10pt}                
        \label{fig:pipeline}
    \end{center}
\end{figure}

A typical machine learning pipeline for inference can be broken down into two steps as shown in Fig.~\ref{fig:pipeline}: Feature Extraction and Classification.  Approaches such as deep neural networks (DNN) blur the distinction between these steps.     

\subsection{Feature Extraction}
Feature extraction is used to transform the raw data into meaningful inputs for the given task.  Traditionally, feature extraction was designed through a \emph{hand-crafted} process by experts in the field.  For instance, for object recognition in computer vision, it was observed that humans are sensitive to edges (i.e., gradients) in an image.  As a result, many well-known computer vision algorithms use image gradient-based features such as Histogram of Oriented Gradients (HOG)~\cite{dalal2005histograms} and Scale Invariant Feature Transform (SIFT)~\cite{lowe1999object}. The challenge in designing these features is to make them robust to variations in illumination and noise.


\subsection{Classification}
The output of feature extraction is represented by a vector, which is mapped to a score using a classifier.  Depending on the application, the score is either compared to a threshold to determine if an object is \emph{present}, or compared to the other scores to determine the object \emph{class}. 

Techniques often used for classification include linear methods such as support vector machine (SVM)~\cite{cristianini2000introduction} and Softmax, and non-linear methods such as  kernel-SVM~\cite{cristianini2000introduction} and Adaboost~\cite{schapire2012boosting}. In many of these classifiers, the computation of the score is effectively a dot product of the features ($\vec{x}$) and the weights ($\vec{w}$) (i.e., $\sum_i w_{i}x_{i}$). As a result, much of the hardware research has been focused on reducing the cost of a multiply and accumulate (MAC).

\subsection{Deep Neural Networks (DNN)}
Rather than using hand-crafted features, the features can be learned directly from the data, similar to the weights in the classifier, such that the entire system is trained end-to-end.  These \emph{learned} features are used in a popular form of machine learning called deep neural networks (DNN), also known as deep learning~\cite{nature2015-lecun}. DNN delivers higher accuracy than hand-crafted features on a variety of tasks~\cite{ijcv2015-russakovsky} by mapping inputs to a high-dimensional space; however, it comes at the cost of high computational complexity. 

\begin{figure}
    \begin{center}
        \includegraphics[width=0.9\linewidth]{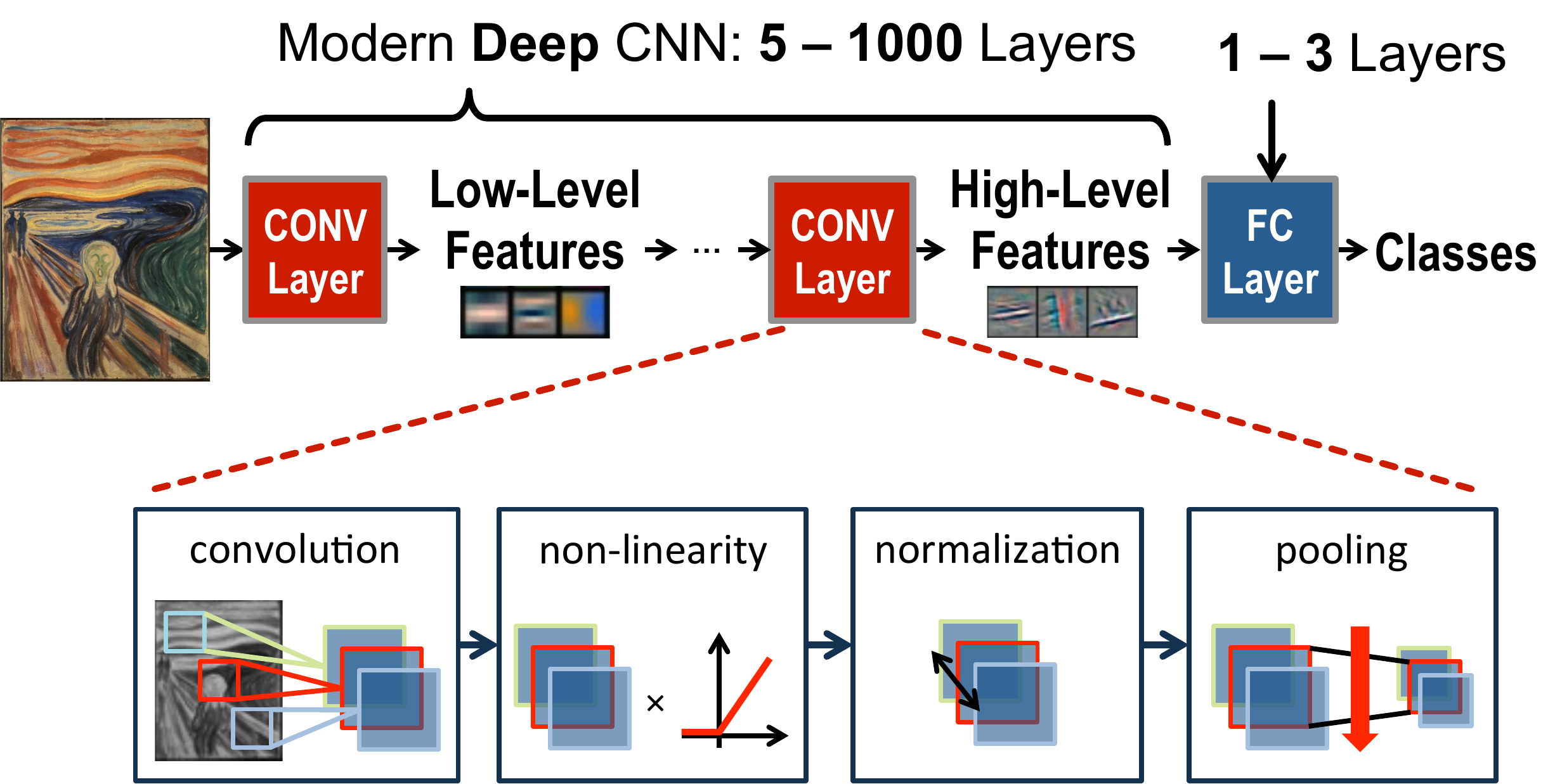}
        \caption{    Deep Neural Networks are composed of several convolutional layers followed by fully connected layers.
                }
        \label{fig:DNN}
    \end{center}
\end{figure}

There are many forms of DNN (e.g., convolutional neural networks, recurrent neural networks, etc.). For computer vision applications, DNNs are composed of multiple convolutional (CONV) layers~\cite{iscas2010-lecun} as shown in Fig.~\ref{fig:DNN}.  With each layer, a higher-level abstraction of the input data, called a feature map, is extracted to preserve essential yet unique information. Modern DNNs are able to achieve superior performance by employing a very deep hierarchy of layers.

\begin{figure}
    \begin{center}
        \includegraphics[width=0.9\linewidth]{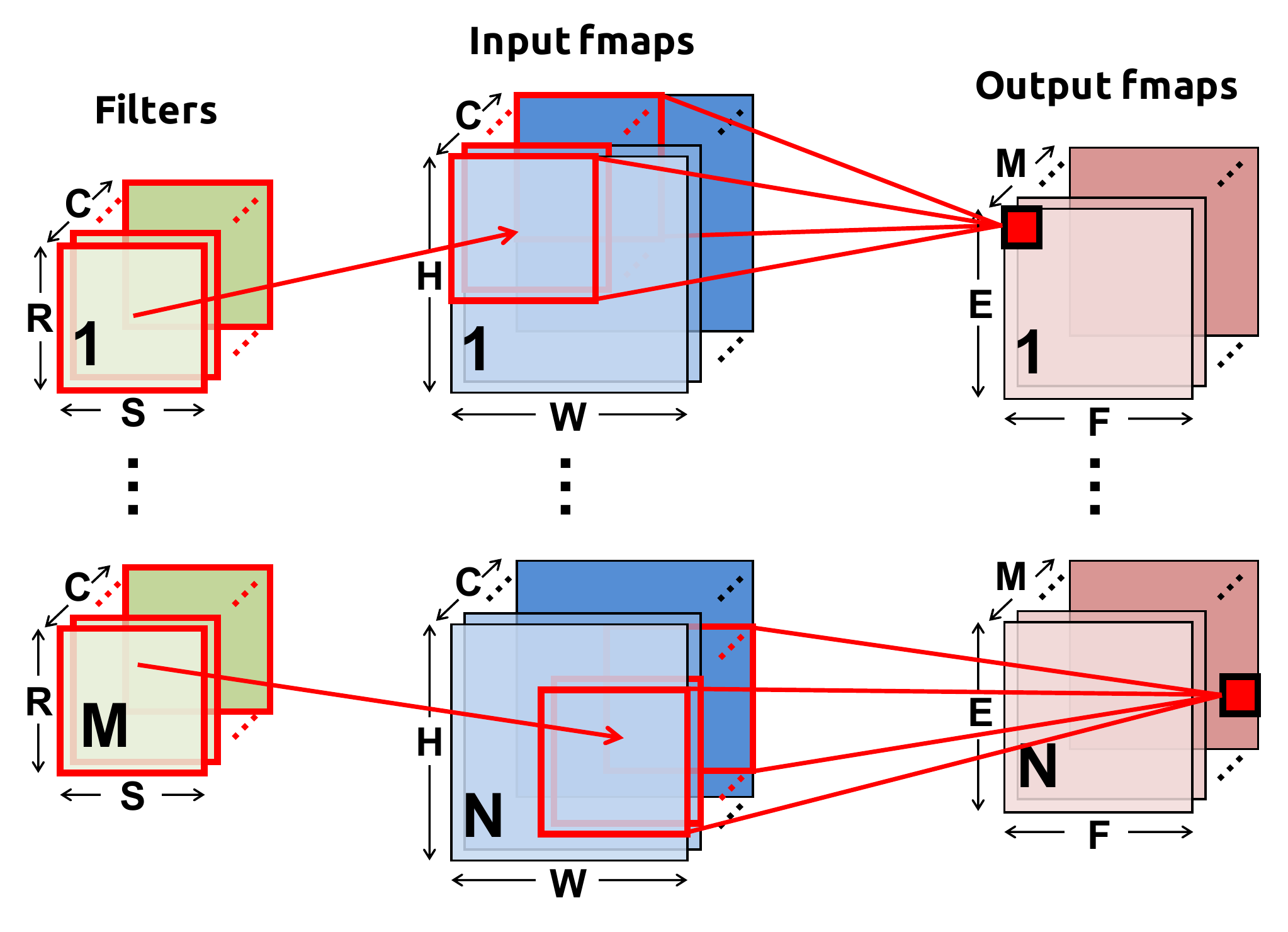}
            \vspace{-10pt}
        \caption{    Computation of a convolution in DNN. 
                }
        \vspace{-10pt}                
        \label{fig:CNN}
    \end{center}
\end{figure}

Fig.~\ref{fig:CNN} shows an example of a convolution in DNNs. The 3-D inputs to each CONV layer are 2-D feature maps ($W\times H$) with multiple channels ($C$). For the first layer, the input would be the 2-D image itself with three channels, typically the red, green and blue color channels. Multiple 3-D filters ($M$ filters with dimension $R\times S \times C$) are then convolved with the input feature maps, and each filter generates a channel in the output 3-D feature map ($E\times F$ with $M$ channels). The same set of $M$ filters is applied to a batch of $N$ input feature maps.  Thus there are $N$ input feature maps and $N$ output feature maps. In addition, a 1-D bias is added to the filtered result.

The output of the final CONV layer is processed by fully-connected (FC) layers for classification purposes.  In FC layers, the filter and input feature map are the same size, so that there is a different weight for each input pixel. The number of FC layers has been reduced from three to one in most recent DNNs~\cite{cvpr2015-szegedy,cvpr2016-he}. In between CONV and FC layers, additional layers can be optionally added, such as the pooling and normalization layers~\cite{ioffe2015batch}. Each of the CONV and FC layers is also immediately followed by an activation layer, such as a rectified linear unit (ReLU)~\cite{icml2014-nair}. Convolutions account for over 90\% of the run-time and energy consumption in DNNs. 


Table~\ref{tab:popular_dnns} compares modern DNNs, with a popular neural net from the 1990s, LeNet-5~\cite{ieee1998-lecun}, in terms of number layers (depth), number of filters weights, and number of operations (i.e., MACs). Today's DNNs are several orders of magnitude larger in terms of compute and storage.  A more detailed discussion on DNNs can be found in~\cite{micro_2016_dnn_tutorial}.


\begin{table}
\centering
\caption{Summary of popular DNNs \cite{ieee1998-lecun,nips2012-krizhevsky,iclr2015-simonyan, cvpr2015-szegedy, cvpr2016-he}.  Accuracy measured based on Top-5 error on ImageNet~\cite{ijcv2015-russakovsky}.}
\begin{tabular}{|c|c|c|c|c|c|}
\hline
\multirow{2}{*}{\textbf{Metrics}} & \textbf{LeNet} & \textbf{AlexNet} & \textbf{VGG}& \textbf{GoogLeNet}& \textbf{ResNet}\\
& \textbf{5} & \textbf{} & \textbf{16}& \textbf{(v1)}& \textbf{50}\\\noalign{\hrule height 2pt}
\textbf{Accuracy}& {n/a} & {16.4} & {7.4}& {6.7}& {5.3}\\\noalign{\hrule height 1.5pt}
\textbf{CONV Layers}& {2} & {5} & {16}& {21}& {49}\\\hline
\textbf{Weights}& {2.6k} & {2.3M} & {14.7M}& {6.0M}& {23.5M}\\\hline
\textbf{MACs}& {283k} & {666M} & {15.3G}& {1.43G}& {3.86G}\\\noalign{\hrule height 1.5pt}
\textbf{FC Layers}& {2} & {3} & {3}& {1}& {1}\\\hline
\textbf{Weights}& {58k} & {58.6M} & {124M}& {1M}& {2M}\\\hline
\textbf{MACs}& {58k} & {58.6M} & {124M}& {1M}& {2M}\\\noalign{\hrule height 1.5pt}
\textbf{Total Weights}& {60k} & {61M} & {138M}& {7M}& {25.5M}\\\hline
\textbf{Total MACs}& {341k} & {724M} & {15.5G}& {1.43G}& {3.9G}\\\hline
\end{tabular}
\label{tab:popular_dnns}
\end{table}

\subsection{Complexity versus Difficulty of Task}
\begin{figure}
    \begin{center}
        \includegraphics[width=0.9\linewidth]{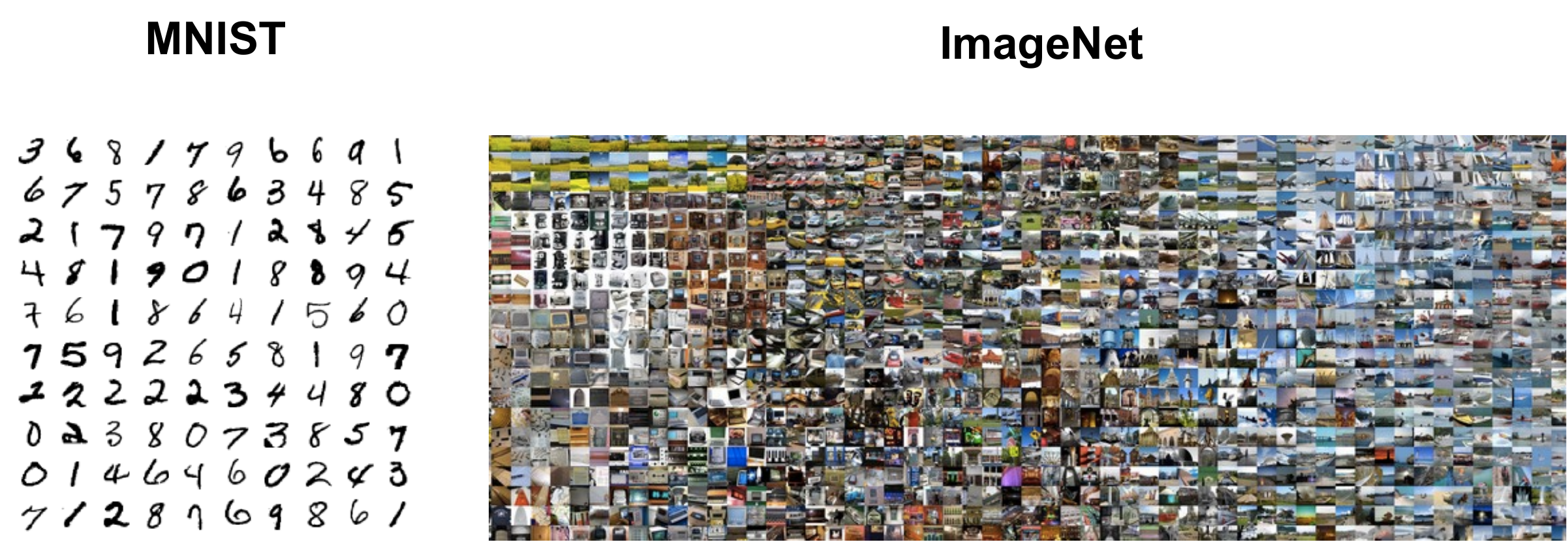}
        \caption{   MNIST (10 classes, 60k training, 10k testing)~\cite{mnist} vs. ImageNet (1000 classes, 1.3M training, 100k testing)\cite{ijcv2015-russakovsky} dataset.
                }
        \label{fig:datasets}
    \end{center}
    \vspace{-10pt}
\end{figure}

It is important to factor in the difficulty of the task when comparing  different machine learning methods.  For instance, the task of classifying handwritten digits from the MNIST dataset~\cite{mnist} is much simpler than classifying an object into one of a 1000 classes as is required for the ImageNet dataset~\cite{ijcv2015-russakovsky}(Fig.~\ref{fig:datasets}).  It is expected that the size of the classifier or network (i.e., number of weights) and the number of MACs will be larger for the more difficult task than the simpler task and thus require more energy.  For instance, LeNet-5\cite{ieee1998-lecun} is designed for digit classification, while AlexNet\cite{nips2012-krizhevsky}, VGG-16\cite{iclr2015-simonyan}, GoogLeNet\cite{cvpr2015-szegedy}, and ResNet\cite{cvpr2016-he} are designed for the 1000 class image classification. 

\section{Challenges} 
The key metrics for embedded machine learning are accuracy, energy consumption, throughput/latency, and cost. 

The accuracy of the machine learning algorithm should be measured on a sufficiently large dataset.  There are many widely-used publicly-available datasets that researcher can use (e.g., ImageNet).


Programmability is important since the weights need to be updated when the environment or application changes. In the case of DNNs, the processor must also be able to support different networks with varying number of layers, filters, channels and filter sizes. 

The high dimensionality and need for programmability both result in an increase in computation and data movement. Higher dimensionality increases the amount of data generated and programmability means that the weights also need be read and stored. This poses a challenge for energy-efficiency since data movement costs more than computation~\cite{isscc2014-horowitz}. In this paper, we will discuss various methods that reduce data movement to minimize energy consumption.  

The throughput is dictated by the amount of computation, which also increases with the dimensionality of the data. In this paper, we will discuss various methods that the data can be transformed to reduce the number of required operations.

The cost is dictated by the amount of storage required on the chip. In this paper, we will discuss various methods to reduce storage costs such that the area of the chip is reduced, while maintaining low off-chip memory bandwidth.

Finally, training requires a significant amount of labeled data (particularly for DNNs) as well as computation for multiple iterations of back-propagation to determine the value of the weights. There is on-going research on training in the cloud using CPUs, GPUs, FPGAs and ASICs. However, this is beyond the scope of this paper.

Currently, state-of-the-art DNNs consume orders of magnitude higher energy than other forms of embedded processing (e.g., video compression). We must exploit opportunities at multiple levels of hardware design to address all these challenges and close this energy gap.

\section{Opportunities in Architectures}
\begin{figure}
    \begin{center}
        \includegraphics[width=0.9\linewidth]{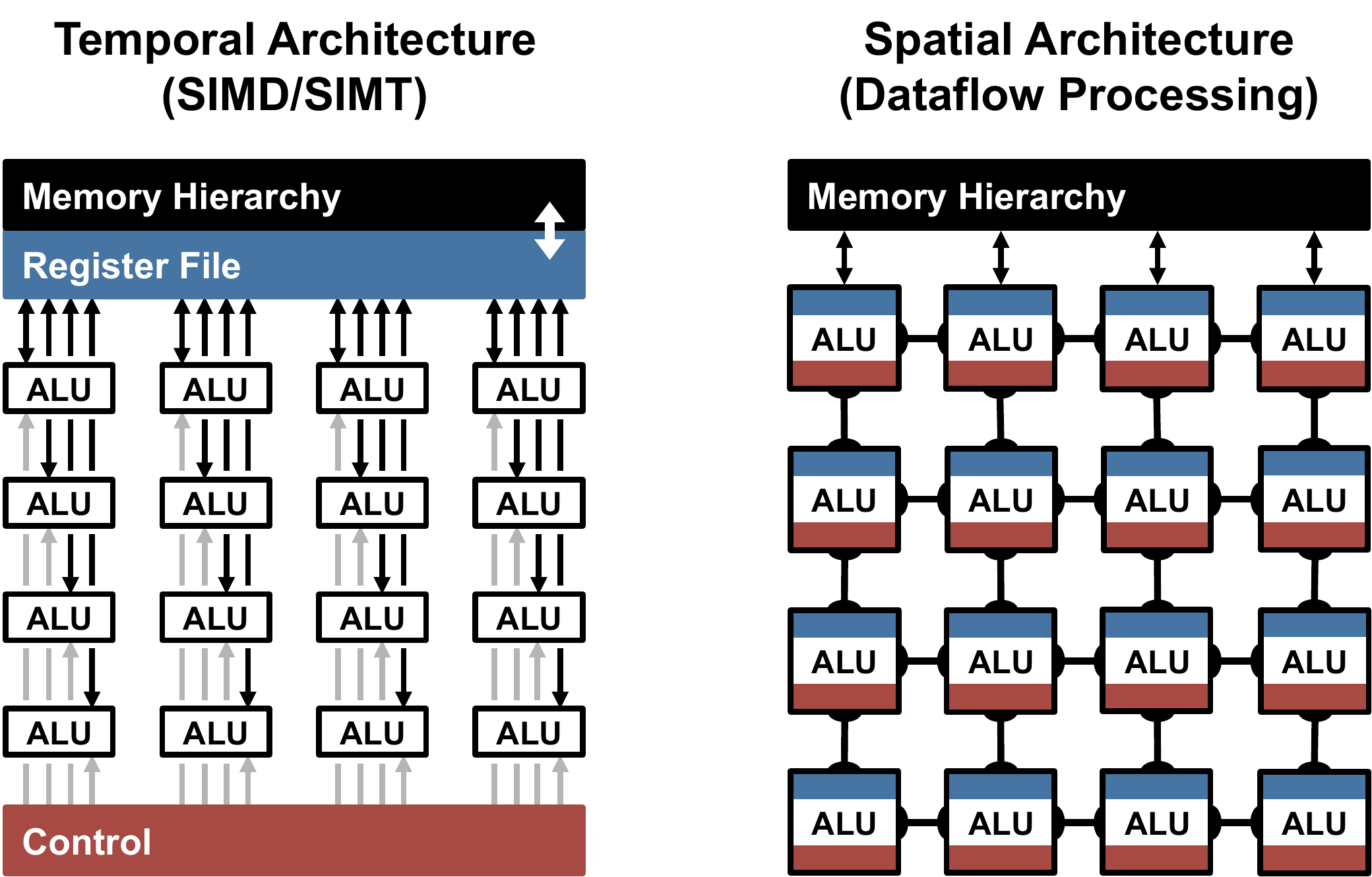}
        \caption{     Highly-parallel compute paradigms.
                }
        \label{fig:parallel_compute}
                \vspace{-10pt}
    \end{center}
\end{figure}

The MAC operations in both the feature extraction (CONV layer in a DNN) and classification (for both DNN and hand-crafted features) can be easily  parallelized. Two common highly-parallel compute paradigms are shown in Fig.~\ref{fig:parallel_compute} with multiple arithmetic logic units (ALU).  

\subsection{CPU and GPU Platforms}
CPUs and GPUs use temporal architectures such as SIMD or SIMT to perform the MACs in parallel.  All the ALUs share the same control and memory (register file).  On these platforms, all classifications are represented by a matrix multiplication. The CONV layer in a DNN can also be mapped to a matrix multiplication using the Toeplitz matrix. There are software libraries designed for CPUs (e.g., OpenBLAS, Intel MKL, etc.) and GPUs (e.g., cuBLAS, cuDNN, etc.) that optimize for matrix multiplications.  The matrix multiplication is tiled to the storage hierarchy of these platforms, which are on the order of a few megabytes at the higher levels.

The matrix multiplications on these platforms can be further sped up by applying transforms to the data to reduce the number of multiplications.  Fast Fourier Transform (FFT)~\cite{mathieu2013fast, dubout2012exact} is a well known approach that reduces the number of multiplications from O($N_{o}^{2}N_{f}^{2}$) to O($N_{o}^{2}log_{2}N_{o})$, where the output size is $N_{o}\times N_{o}$ and the filter size is $N_{f}\times N_{f}$; however, the benefits of FFTs decrease with filter size.  Other approaches include Strassen~\cite{cong2014minimizing} and Winograd~\cite{lavin2015fast}, which rearrange the computation such that the number of multiplications scale from O($N^{3}$) to O($N^{2.807}$) and 2.25$\times$ for a $3\times 3$ filter, respectively, at the cost of reduced numerical stability, increased storage requirements, and specialized processing depending on the size of the filter.

\subsection{Accelerators}
\begin{figure}
    \begin{center}
        \includegraphics[width=0.9\linewidth]{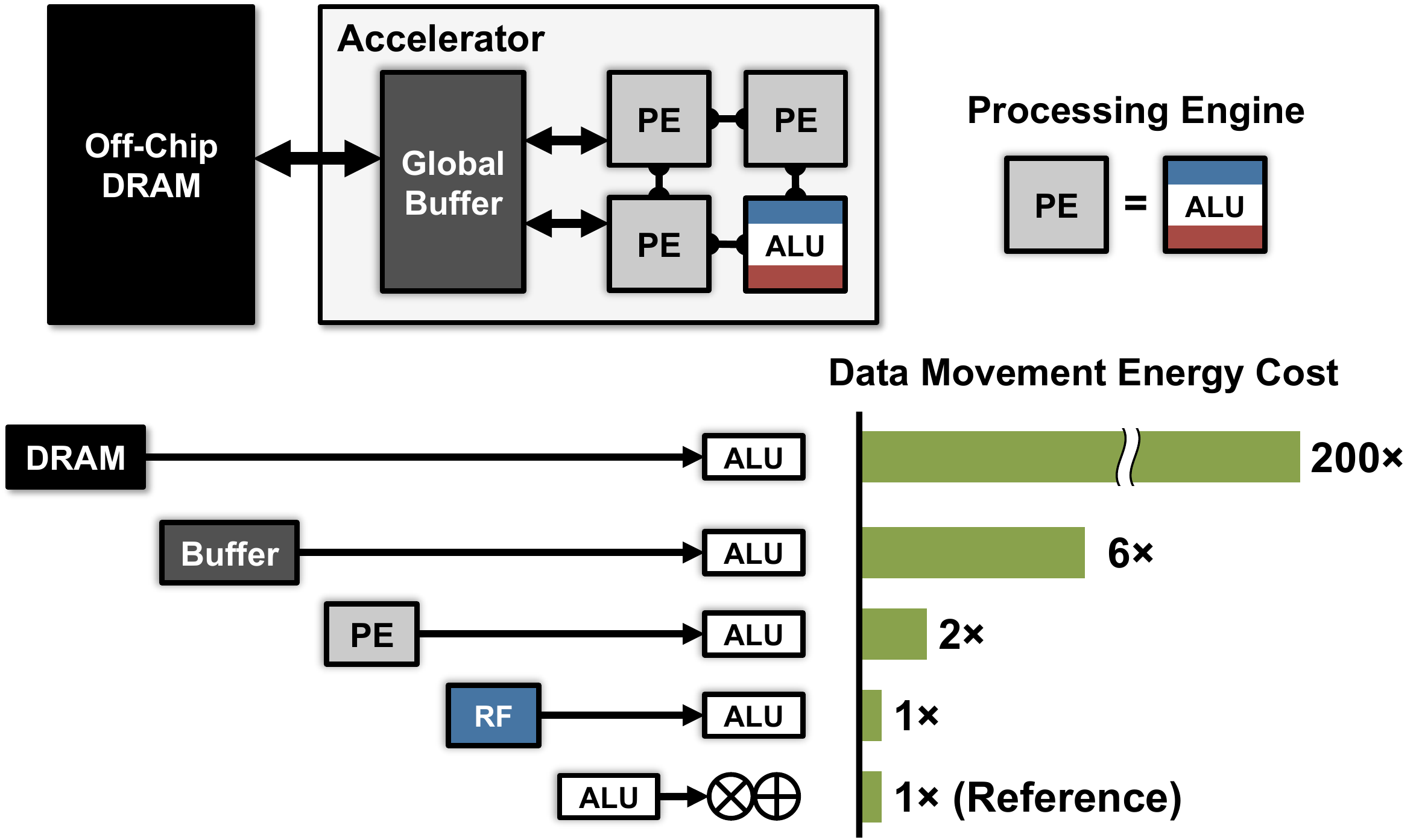}
        \caption{    Memory hierarchy and data movement energy~\cite{isca2016-chen}.
                }      
        \label{fig:data_movement_energy}
    \end{center}
    \vspace{-10pt}
\end{figure}

Accelerators provide an opportunity to optimize the data movement (i.e., dataflow) in order to minimize accesses from the expensive levels of the memory hierarchy as shown in Fig.~\ref{fig:data_movement_energy}. In particular, for DNNs we investigate dataflows that exploit three forms of data reuse (convolutional, filter and image). We use a spatial architecture (Fig.~\ref{fig:parallel_compute}) with local memory (register file) at each ALU processing element (PE) on the order of 0.5 -- 1.0kB and a shared memory (global buffer) on the order of 100 -- 500kB. The global buffer communicates with the off-chip memory (e.g., DRAM).  Data movement is allowed between the PEs using an on-chip network (NoC) to reduce accesses to the global buffer and the off-chip memory.  Three types of data movement include input pixels, filter weights and partial sums (i.e., the product of pixels and weights) that are accumulated for the output.

Recent work~\cite{asap2009-sankaradas, fpt2010-sriram, isca2010-chakradhar, cvprw2014-gokhale, isscc2015-park, glsvlsi2015-cavigelli, arxiv2015-gupta, isca2015-du, iccd2013-peemen, fpga2015-zhang, asplos2014-chen, micro2014-chen} has proposed solutions for DNN acceleration, but it is difficult to compare their performance directly due to differences in implementation and design choices. The following taxonomy (Fig.~\ref{fig:dataflow}) can be used to classify these existing DNN dataflows based on their data handling characteristics~\cite{isca2016-chen}: 

\begin{figure}
\centering{
    \subfigure[Weight Stationary]{
		\includegraphics[width=0.9\linewidth]{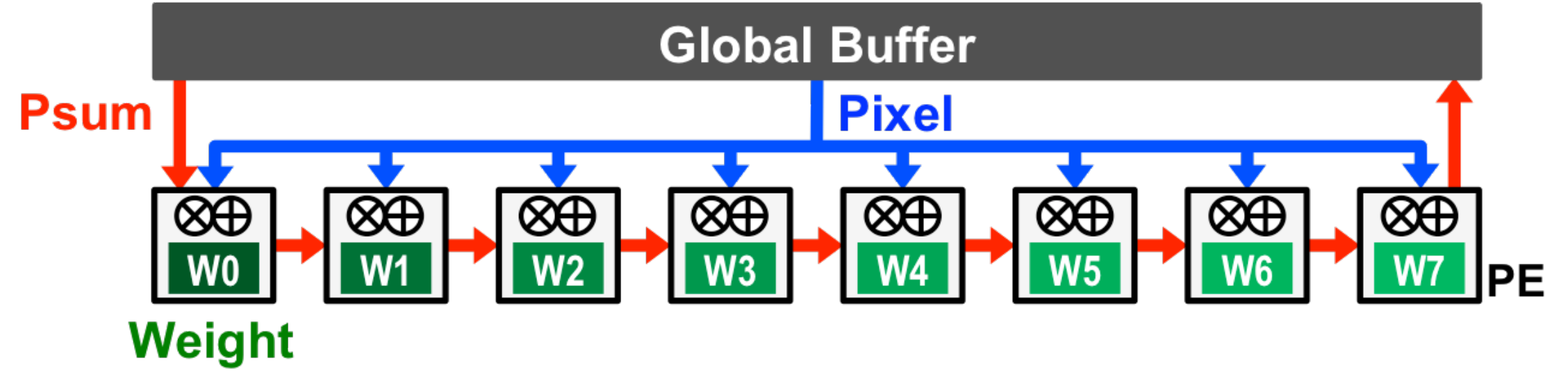}    
		\label{fig:weight_stationary}
	}	
    \subfigure[Output Stationary]{
		\includegraphics[width=0.9\linewidth]{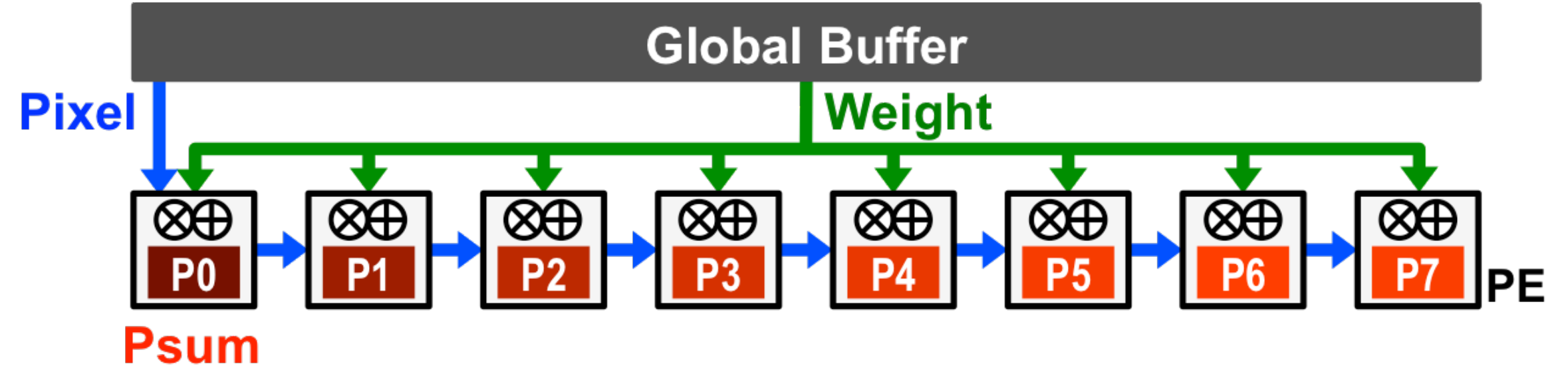}
				\label{fig:output_stationary}
	}
    \subfigure[No Local Reuse]{
		\includegraphics[width=0.9\linewidth]{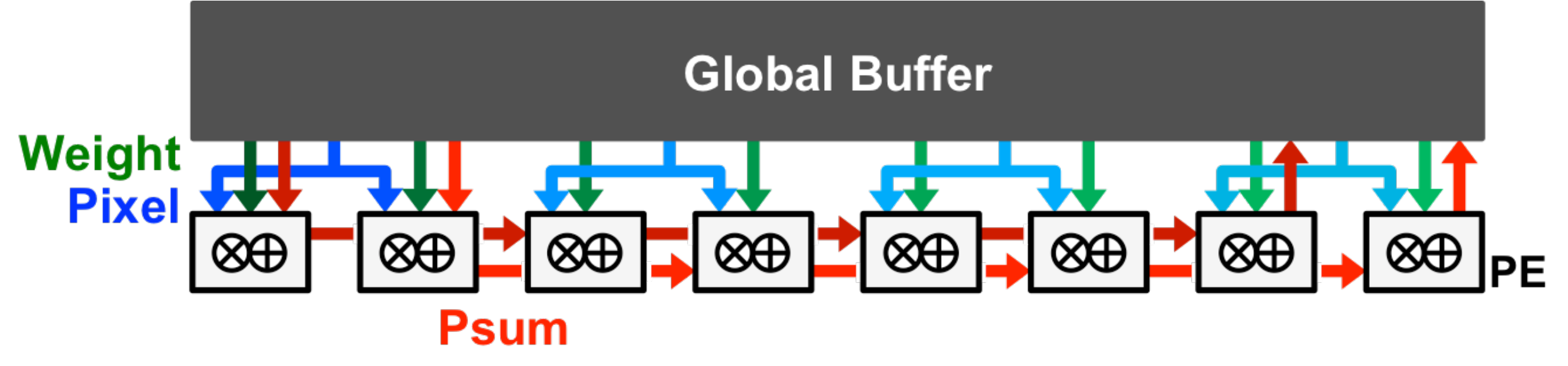}
				\label{fig:no_local_reuse}
	}	
	\vspace{-5pt}
}
\caption{Dataflows for DNNs.}
            
\label{fig:dataflow}
\end{figure}

\begin{figure}
    \begin{center}
        \includegraphics[width=0.9\linewidth]{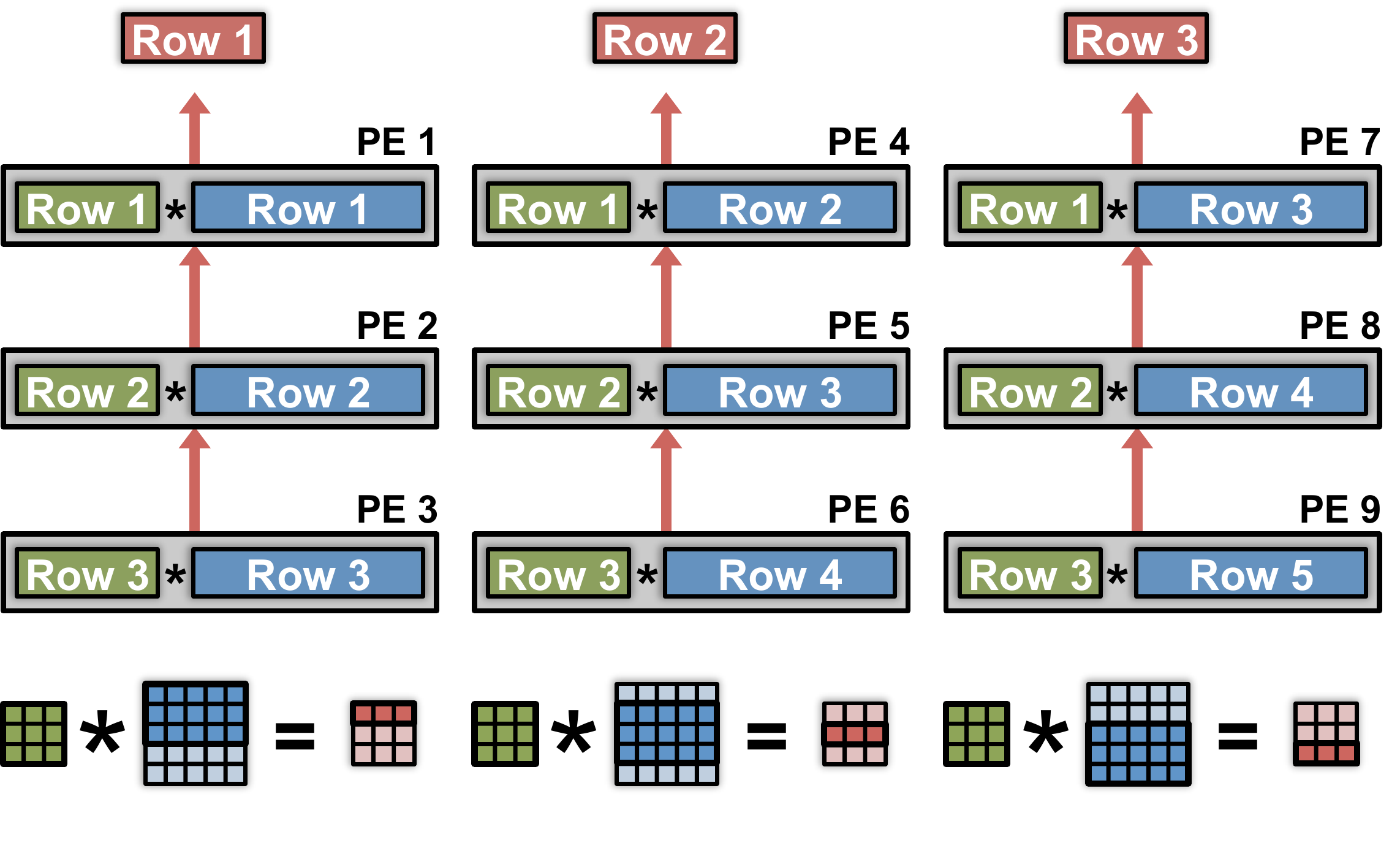}
                   \vspace{-10pt} 
        \caption{    Row Stationary Dataflow~\cite{isca2016-chen}.
                }
                  \vspace{-10pt} 
        \label{fig:row_stationary}
    \end{center}
\end{figure}

\begin{itemize}
\item \textbf{Weight stationary (WS):}
The weights are stored in the register file at the PE and remains stationary to minimized the movement cost of the weights (Fig.~\ref{fig:weight_stationary}). The inputs and partial sums must move through the spatial array and global buffer. Examples are found in~\cite{asap2009-sankaradas, fpt2010-sriram, isca2010-chakradhar, cvprw2014-gokhale, isscc2015-park, glsvlsi2015-cavigelli}.

\item \textbf{Output stationary (OS):}
The outputs are stored in the register file at the PE and remains stationary to minimized the movement cost of the partial sums (Fig.~\ref{fig:output_stationary}). The inputs and weights must move through the spatial array and global buffer. Examples are found in~\cite{arxiv2015-gupta, isca2015-du, iccd2013-peemen}.

\item \textbf{No local reuse (NLR):}
While small register files are efficient in terms of energy (pJ/bit), they are inefficient in terms area ($\mu m^{2}$/bit).  In order to maximize the storage capacity, and minimize the off-chip memory bandwidth, no local storage is allocated to the PE and instead all that area is allocated to the global buffer to increase its capacity (Fig.~\ref{fig:no_local_reuse}).  The trade-off is that there will be increased traffic on the spatial array and to the global buffer for all data types. Examples are found in~\cite{fpga2015-zhang, asplos2014-chen, micro2014-chen}.

\item \textbf{Row stationary (RS):}
In order to increase reuse of all types of data (weights, pixels, partial sums), a row stationary approach is proposed in~\cite{isca2016-chen}.  A row of the filter convolution remains stationary within a PE to exploit 1-D convolutional reuse within the PE. Multiple 1-D rows are combined in the spatial array to exhaustively exploit all convolutional reuse (Fig.~\ref{fig:row_stationary}), which reduces accesses to the global buffer.  Multiple 1-D rows from different channels and filters are mapped to each PE to reduce partial sum data movement and exploit filter reuse, respectively.  Finally, multiple passes across the spatial array allow for additional image and filter reuse using the global buffer. This dataflow is demonstrated in~\cite{isscc2016-chen}.
\end{itemize}

\begin{figure}
\centering{
    \subfigure[Across types of data]{
		\includegraphics[width=0.9\linewidth]{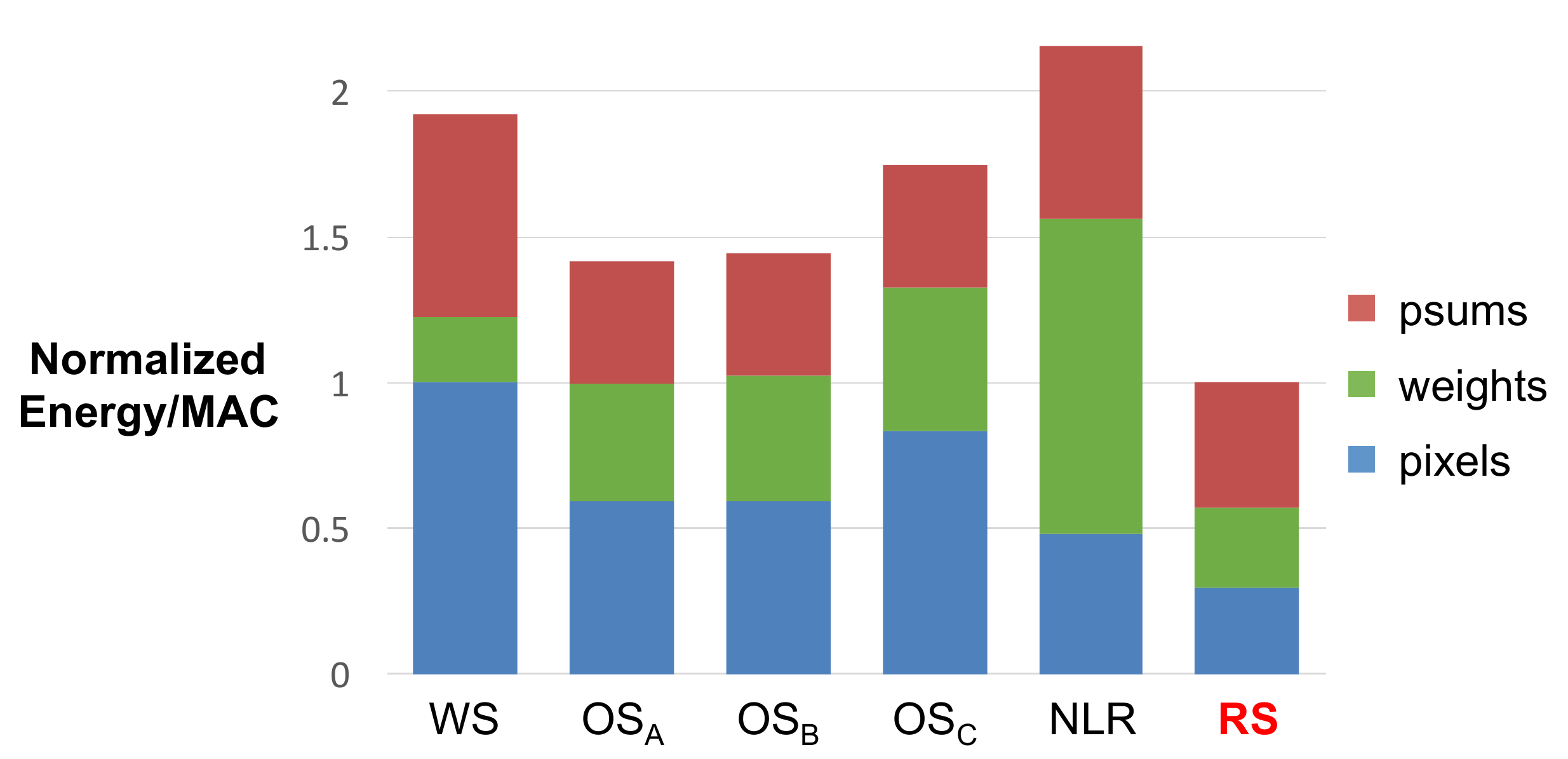}
		\label{fig:energy_DNN_dataflow_type}
	}	
	\vspace{10pt}
    \subfigure[Across levels of memory hierarchy]{
		\includegraphics[width=0.9\linewidth]{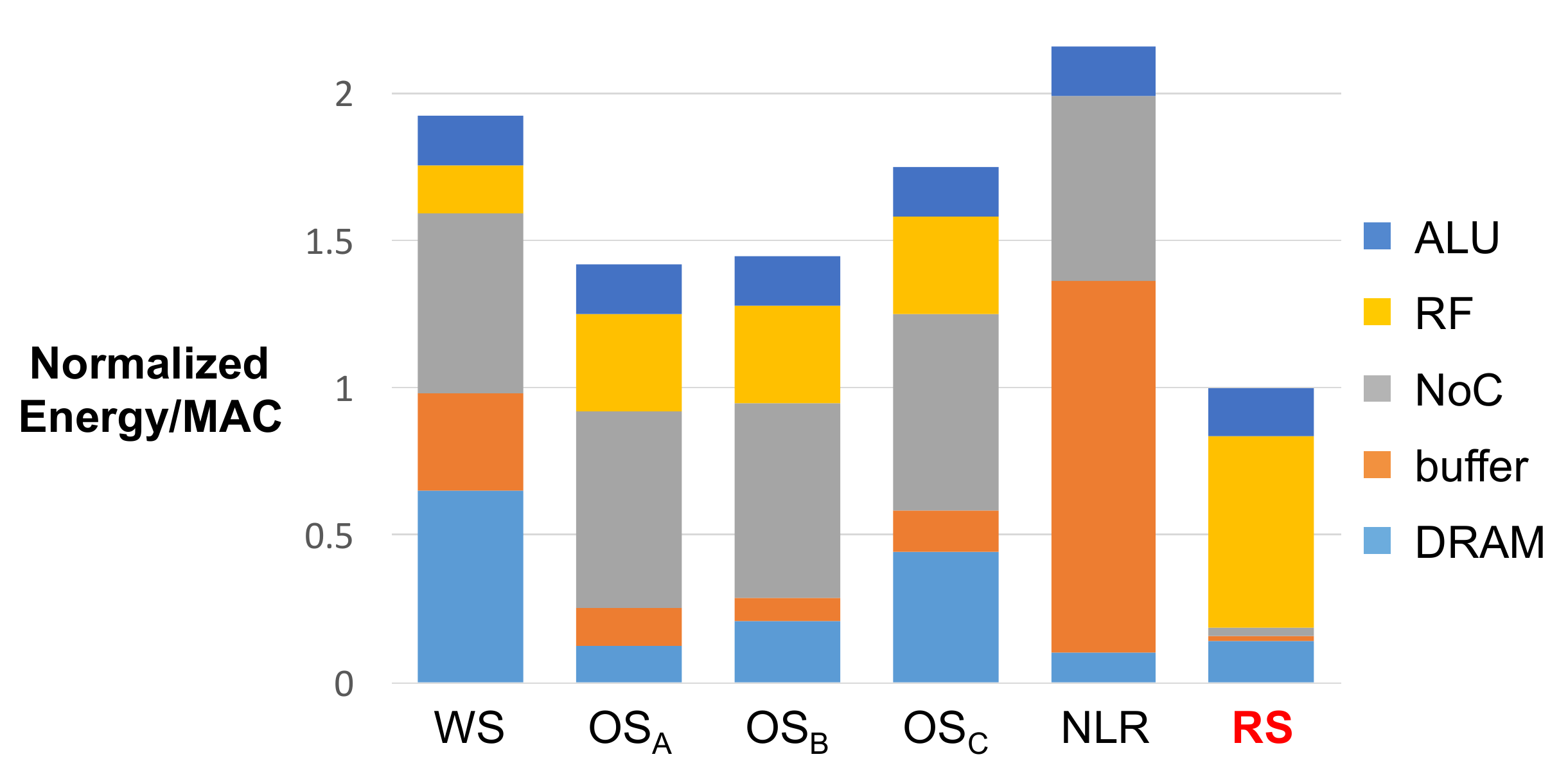}
				\label{fig:energy_DNN_dataflow_mem}
	}
}
	\vspace{-10pt}
\caption{Energy breakdown of dataflows~\cite{isca2016-chen}.}
\label{fig:dataflow_energy_comparison}
\end{figure}

The dataflows are compared on a spatial array with the same number of PEs (256), area cost and DNN (AlexNet). Fig.~\ref{fig:dataflow_energy_comparison} shows the energy consumption of each approach. The row stationary approach is 1.4$\times$ to 2.5$\times$ more energy-efficient than the other dataflows for the convolutional layers.  This is due to the fact that the energy of all types of data is reduced.  Furthermore, both the on-chip and off-chip energy is considered.

\section{Opportunities in Joint Algorithm and Hardware Design} 
\label{sec:algorithm}
There is on-going research on modifying the machine learning algorithms to make them more hardware-friendly while maintaining  accuracy; specifically, the focus is on reducing computation, data movement and storage requirements.  

\subsection{Reduce Precision}
The default size for programmable platforms such as CPUs and GPUs is often 32 or 64 bits with floating-point representation. While this remains the case for training, during inference, it is possible to use a fixed-point representation and substantially reduce the bitwidth for energy and area savings, and increase in throughput.  Retraining is typically required to maintain accuracy when pushing the weights and features to lower bitwidth.

In hand-crafted approaches, the bitwidth can be drastically reduced to below 16-bits without impacting the accuracy. For instance, in object detection using HOG, each 36-dimension feature vector only requires 9-bit per dimension, and each weight of the SVM uses only 4-bits~\cite{suleiman2014energy}; for object detection using deformable parts models (DPM)~\cite{felzenszwalb2010object}, only 11-bits are required per feature vector and only 5-bits are required per SVM weight~\cite{suleiman201658}.

Similarly for DNN inference, it is common to see accelerators support 16-bit fixed point~\cite{isscc2016-chen, asplos2014-chen}.  There has been significant research on exploring the impact of bitwidth on accuracy~\cite{gysel2016hardware}. In fact, recently commercial hardware for DNN reportedly support 8-bit integer operations~\cite{tpu}.  As bitwidths can vary by layer, hardware optimizations have been explored to exploit the reduced bitwidth for 2.56$\times$ energy savings~\cite{moons20160} or 2.24$\times$ increase in throughput~\cite{judd2016stripes} compared to a 16-bit fixed point implementation. With more significant changes to the network, it is possible to reduce bitwidth down to 1-bit for either weights~\cite{nips2015-courbariaux-binaryconnect} or both weights and activations~\cite{courbariaux2016binarynet,eccv2016-rastegari-xnor_net} at the cost of reduced accuracy. The impact of 1-bit weights on hardware is explored in~\cite{andri2016yodann}.

\subsection{Sparsity}
\begin{figure}
    \begin{center}
        \includegraphics[width=0.9\linewidth]{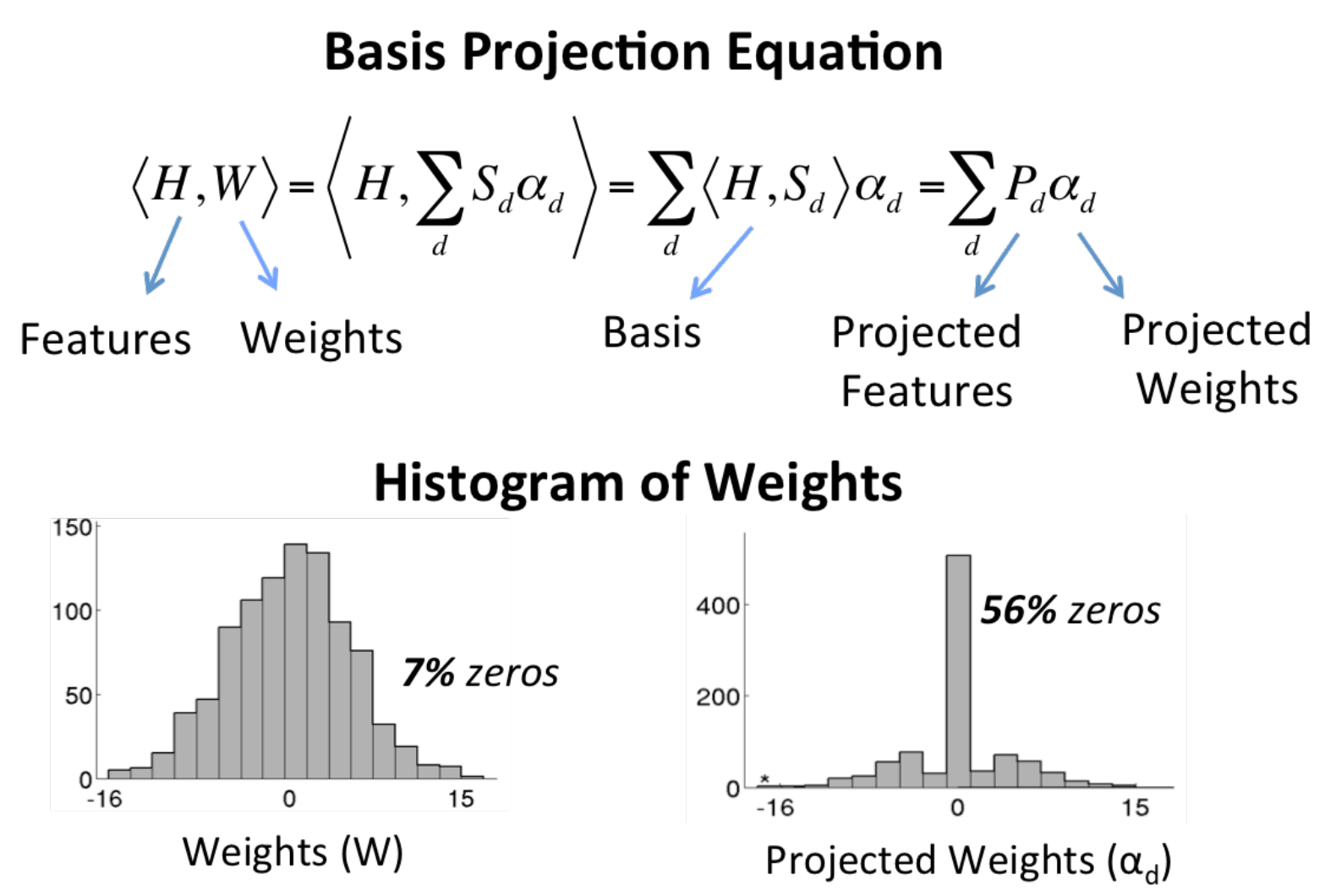}
                \vspace{-5pt}
        \caption{    Sparse weights after basis projection~\cite{suleiman201658}.
                }
        \label{fig:sparse}
        \vspace{-10pt}
    \end{center}
\end{figure}

For SVM classification, the weights can be projected onto a basis such that the resulting weights are sparse for a 2$\times$ reduction in number of multiplications~\cite{suleiman201658} (Fig.~\ref{fig:sparse}).  For feature extraction, the input image can be made sparse by pre-processing for a 24\% reduction in power consumption~\cite{suleiman2014energy}. 

For DNNs, the number of MACs and weights can be reduced by removing weights through a process called pruning.  This was first explored in \cite{nips1990-lecun-opt_brain_damage} where weights with minimal impact on the output were removed.  In~\cite{nips2015-han}, pruning is applied to modern DNNs by removing small weights.  However, removing weights does not necessarily lead to lower energy.  Accordingly, in~\cite{yang2016} weights are removed based on an energy-model to directly minimize energy consumption.  The tool used for energy modeling can be found at~\cite{energy_estimation}.

Specialized hardware has been proposed in~\cite{isscc2016-chen,suleiman201658,albericio2016cnvlutin, han2016eie} to exploit sparse weights for increased speed or reduced energy consumption. In Eyeriss~\cite{isscc2016-chen}, the processing elements are designed to skip reads and MACs when the inputs are zero, resulting in a 45\% energy reduction.  In~\cite{suleiman201658}, by using specialized hardware to avoid sparse weights, the energy and storage cost are reduced by 43\% and 34\%, respectively. 

\subsection{Compression}
Data movement and storage are important factors in both energy and cost. Feature extraction can result in sparse data (e.g., gradient in HOG and ReLU in DNN) and the weights used in classification can also be made sparse by pruning. As a result, compression can be applied to exploit data statistics to reduce data movement and storage cost.

Various forms of lightweight compression have been explored to reduce data movement. Lossless compression can be used to reduce the transfer of data on and off chip~\cite{chen20101,moons20160,han2016eie}. Simple run-length coding of the activations in~\cite{jssc2017-chen} provides up to 1.9$\times$ bandwidth reduction, which is within 5-10\% of the theoretical entropy limit. Lossy compression such as vector quantization can also be used on feature vectors~\cite{suleiman201658} and weights~\cite{iclr2016-han, lee2013low, price20156} such that they can be stored on-chip at low cost. Generally, the cost of the compression/decompression is on the order of a few thousand kgates with minimal energy overhead. In the lossy compression case, it is also important to evaluate the impact on performance accuracy.

\section{Opportunities in Mixed-Signal Circuits}
Most of the data movement is in between the memory and processing element (PE), and also the sensor and PE.  In this section, we discuss how this is addressed using mixed-signal circuit design.  However, circuit non-idealities should also be factored into the algorithm design; these circuits can benefit from the reduced precision algorithms discussed in Section~\ref{sec:algorithm}. In addition, since the training often occurs in the digital domain, the ADC and DAC conversion overhead should also be accounted for when evaluating the system. 

While spatial architectures bring the memory closer to the computation (i.e., into the PE), there have also been efforts to integrate the computation into the memory itself.  For instance, in~\cite{zhang2016machine} the classification is embedded in the SRAM.  Specifically, the word line (WL) is driven by a 5-bit feature vector using a DAC, while the bit-cells store the binary weights $\pm 1$.  The bit-cell current is effectively a product of the value of the feature vector and the value of the weight stored in the bit-cell; the currents from the column are added together to discharge the bitline (BL or BLB). A comparator is then used to compare the resulting dot product to a threshold, specifically sign thresholding of the differential bitlines. Due to the variations in the bitcell, this is considered a weak classifier, and boosting is needed to combine the weak classifiers to form a strong classifier~\cite{wang2014error}. This approach gives 12$\times$ energy savings over reading the 1-bit weights from the SRAM.

Recent work has also explored the use of mixed-signal circuits to reduce the computation cost of the MAC. It was shown in~\cite{murmann2015mixed} that performing the MAC using switched capacitors can be more energy-efficient than digital circuits despite ADC and DAC conversion overhead. Accordingly, the matrix multiplication can be integrated into the ADC as demonstrated in~\cite{zhang201518}, where the most significant bits of the multiplications for Adaboost classification are performed using switched capacitors in an 8-bit successive approximation format.  This is extended in~\cite{lee201624} to not only perform multiplications, but also the accumulation in the analog domain. It is assumed that 3-bits and 6-bits are sufficient to represent the weights and input vectors, respectively. This enables the computation to move closer to the sensor and reduces the number of ADC conversions by 21$\times$. 

To further reduce the data movement from the sensor,~\cite{likamwa2016redeye} proposed performing the entire convolution layer (including convolution, max pooling and quantization) in the analog domain at the sensor. Similarly, in~\cite{choi20143}, the entire HOG feature is computed in the analog domain to reduce the sensor bandwidth by 96.5\%.

\section{Opportunities in Advanced Technologies} 
In the previous section, we discussed how data movement can be reduced by moving the processing near the memory or the sensor using mixed-signal circuits.  In this section, we will discuss how this can be achieved with advanced technologies.

The use of advanced memory technologies such as embedded DRAM (eDRAM) and Hybrid Memory Cube (HMC) are explored in~\cite{micro2014-chen} and~\cite{kim2016neurocube}, respectively, to reduce the energy access cost of the weights in DNN.  There has also been a lot of work that investigates integrating the multiplication directly into advanced \emph{non-volatile} memories by using them as resistive elements.  Specifically, the multiplications are performed where the conductance is the weight, the voltage is the input, and the current is the output (note: this is the ultimate form of weight stationary, as the weights are always held in place); the addition is done by summing the current using Kirchhoff's current law. In~\cite{shafiee2016isaac}, memristors are used to compute a 16-bit dot product operation with 8 memristors each storing 2-bits; a 1-bit$\times$2-bit multiplication is performed at each memristor, where a 16-bit input requires 16 cycles to complete.  In~\cite{chi2016prime}, ReRAM is used to compute the product of a 3-bit input and 4-bit weight.  Similar to the mixed-signal circuits, the precision is limited, and the ADC and DAC conversion overhead must be considered in the overall cost, especially when the weights are trained in the digital domain.  The conversion overhead can be avoided by training directly in the analog domain as shown for the fabricated memristor array in~\cite{prezioso2015training}.

Finally, it may be feasible to embed the computation into the sensor itself. This is useful for image processing where the bandwidth to read the data from the sensor accounts for a significant portion of the system energy consumption.  For instance, an Angle Sensitive Pixels sensor can be used to compute the gradient of the input, which along with compression, reduces the sensor bandwidth by 10$\times$~\cite{wang2012180nm}. A sensor that outputs gradients can also reduce the computation and energy consumption of subsequent processing engine~\cite{chen2016asp,suleiman2014energy}.

\section{Hand-crafted versus Learned Features}
\begin{figure}
    \begin{center}
        \includegraphics[width=0.9\linewidth]{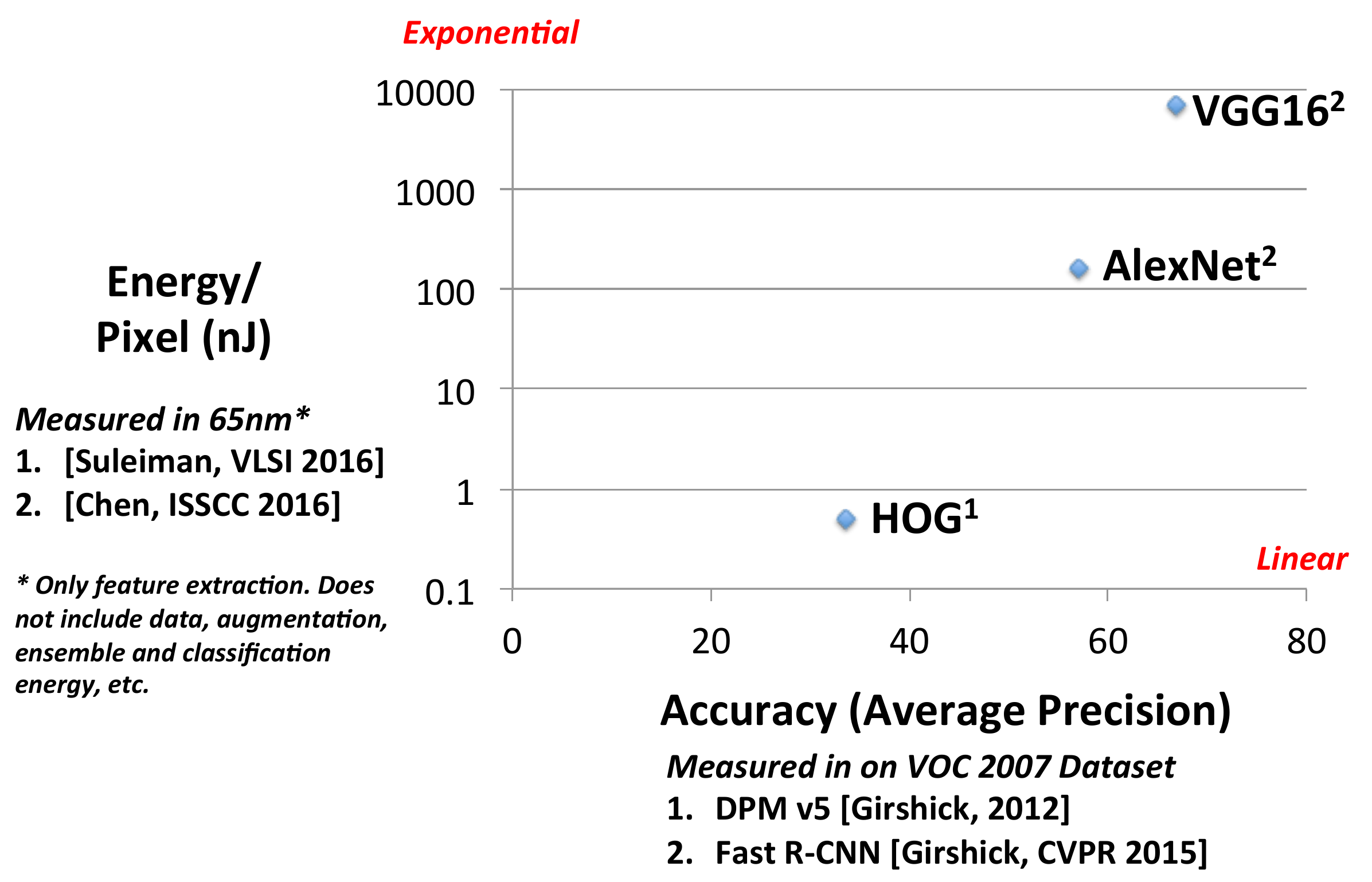}
        \vspace{-5pt}          
        \caption{    Energy vs.\ accuracy comparison of hand-crafted and learned features.
                }
        \vspace{-15pt}                
        \label{fig:hand_vs_dnn}
    \end{center}
\end{figure}

Hand-crafted approaches give higher energy efficiency at the cost of reduced accuracy as compared with learned features such as DNNs.  For hand-crafted features, the amount of computation is less, and reduced bit-width is supported. Furthermore, less data movement is required since the weights are not required for the features.  The classification weights for both approaches must however remain programmable.  Fig.~\ref{fig:hand_vs_dnn} compares the energy consumption of HOG feature extraction versus the convolution layers in AlexNet and VGG-16 based measured results from fabricated 65nm chips~\cite{suleiman201658} and~\cite{isscc2016-chen}, respectively. Note that HOG feature extraction consumes around the same energy as video compression (under 1nJ/pixel~\cite{lin20130} for real-time high definition video), which servers as a good benchmark of what is acceptable for energy consumption near the sensor; however, DNNs currently consume several orders of magnitude more. A more detailed comparison can be found in~\cite{iscas_2017}. We hope that the many design opportunities that we have highlighted in this paper will help close this gap.

\section{Summary} 
Machine learning is an important area of research with many promising applications and opportunities for innovation at various levels of hardware design.  During the design process, it is important to balance the accuracy, energy, throughput and cost requirements.

Since data movement dominates energy consumption, the primary focus of recent research has been to reduce the data movement while maintaining accuracy, throughput and cost.  This means selecting architectures with favorable memory hierarchies like a spatial array, and developing dataflows that increase data reuse at the low-cost levels of the memory hierarchy. With joint design of algorithm and hardware, reduced bitwidth precision, increased sparsity and compression are used to minimize the data movement requirements.  With  mixed-signal circuit design and advanced technologies, computation is moved closer to the source by embedding computation near or within the sensor and the memories.  

One should also consider the interactions between these different levels.  For instance, reducing the bitwidth through hardware-friendly algorithm design enables reduced precision processing with mixed-signal circuits and non-volatile memory. Reducing the cost of memory access with advanced technologies could result in more energy-efficient dataflows.

\section*{Acknowledgment}
Funding provided by DARPA YFA, MIT CICS, TSMC University Shuttle, and gifts from Texas Instruments and Intel.



%


\bibliographystyle{IEEEtran}
\small
\bibliography{IEEEabrv,references}

\end{document}